\documentclass[runningheads]{llncs}

\usepackage{eccv}

\usepackage{eccvabbrv}

\usepackage{graphicx}
\usepackage{booktabs}
\usepackage{multirow}
\usepackage[accsupp]{axessibility}  % Improves PDF readability for those with disabilities.
\usepackage{subcaption}
\usepackage{caption}
\usepackage{makecell}
\usepackage{wrapfig}
\usepackage{algorithm}
\usepackage{algpseudocode}
\usepackage{amsmath}
\usepackage{setspace}
\usepackage[subtle]{savetrees}
\usepackage{hyperref}

\usepackage{orcidlink}

\begin{document}

% ---------------------------------------------------------------
% TODO REVIEW: Replace with your title
\title{Recurrent Autoregressive Diffusion:\\ Global Memory Meets Local Attention} 

% TODO REVIEW: If the paper title is too long for the running head, you can set
% an abbreviated paper title here. If not, comment out.
% \titlerunning{RAD}

% TODO FINAL: Replace with your author list. 
% Include the authors' OCRID for the camera-ready version, if at all possible.

\author{Taiye Chen\inst{1}\orcidlink{0009-0000-8529-8384} \and
Zihan Ding\inst{2} \and
Anjian Li\inst{2} \and
Christina Zhang\inst{2} \and \\
Zeqi Xiao\inst{3} \and
Yisen Wang\inst{1} \and
Chi Jin\inst{2}}

% TODO FINAL: Replace with an abbreviated list of authors.
\authorrunning{T. Chen, Z. Ding, A. Li, C. Zhang, Z. Xiao, Y. Wang, C. Jin}
% First names are abbreviated in the running head.
% If there are more than two authors, 'et al.' is used.

% TODO FINAL: Replace with your institution list.
\institute{
Peking University, China
\email{yeyutaihan@stu.pku.edu.cn,yisen.wang@pku.edu.cn}
\and
Princeton University, USA
\email{\{zihand,anjianl,christinaz,chij\}@princeton.edu}
\and
Nanyang Technological University, Singapore
\email{zeqixiao1@gmail.com}
}

\maketitle

\begin{abstract}
% Recent advancements in video generation have demonstrated the potential of using video diffusion models as world models, with autoregressive generation of ultra long videos through masked conditioning.
Recent advancements in video generation has shifted from bidirectional models for short videos to autoregressive ones for ultra long video generation. 
Previous models, which usually use sliding window attention to restrict inference cost, lack effective memory compression and retrieval for long-term generation beyond the window size, leading to issues of forgetting and spatiotemporal inconsistencies.
% To enhance the retention of historical information within a fixed memory budget, we introduce a recurrent neural network (RNN) into the diffusion transformer framework. Specifically, a diffusion model incorporating LSTM with attention achieves comparable performance to state-of-the-art RNN blocks, such as Test-Time Training (TTT) and Mamba2. 
To enhance the retention of historical information with a fixed memory budget, we additionally incorporate temporal recurrent neural network (RNN) layers into the diffusion transformer (DiT) model. Specifically, we found that a LSTM layer after attention at each DiT layer achieves comparable performance to other state-of-the-art RNN blocks, such as Test-Time Training (TTT) and Mamba2. 
Moreover, existing diffusion-RNN approaches often suffer from performance degradation due to training-inference gap or the lack of overlap across windows. To address these limitations, we propose a novel Recurrent Autoregressive Diffusion (RAD) framework, which leverages recurrent blocks for memory update and retrieval and preserves local details by full attention on overlapping sliding windows, with no training and inference gap. Experiments on Memory Maze and Minecraft datasets demonstrate the superiority for long video generation by our framework with global memory and local attention.
% highlighting the efficiency of LSTM in sequence modeling.

Project page: \href{https://yeyutaihan.github.io/recurrent-autoregressive-diffusion/}{https://yeyutaihan.github.io/recurrent-autoregressive-diffusion/}
\end{abstract}
\section{Introduction}

World models have attracted considerable interest from the research community for their pivotal roles in data synthesis, model-based planning, simulation, and beyond. Leveraging the recent great progress in generative models, video diffusion models have become one of the most promising approaches for efficient and scalable world models. Opposed to representation world models \cite{bardes2024revisitingfeaturepredictionlearning, assran2025vjepa2selfsupervisedvideo} that learn latent representations of the environment, video diffusion models \cite{videoworldsimulators2024, parkerholder2024genie2, genie3} directly achieve world modeling in high-dimensional video pixel space.

Despite remarkable advancements, existing video world models face significant challenges in maintaining spatiotemporal consistency. For instance, models like Oasis \cite{oasis2024}—centered on Minecraft gameplay—and foundational models such as Cosmos \cite{nvidia2025cosmosworldfoundationmodel} both struggle with severe forgetting issues. This difficulty primarily stems from the inherently limited attention window of diffusion transformer (DiT) \cite{peebles2023scalable} architecture. Given the low information density of video data, tokenizing video sequences often results in context lengths that quickly exceed these attention limits. Consequently, frames outside the active attention window are effectively disregarded, resulting in visible temporal and spatial inconsistencies. While recent research has explored alternative approaches to mitigate this, such as leveraging 3D representations \cite{marble2025} to bolster spatial-temporal consistency, they usually lack the interactiveness and scalability of pixel-based video diffusion models.
% Moreover, 3D-based methods typically require high-quality 3D datasets or auxiliary 3D models, which hinders their scalability for large-scale or diverse training scenarios.

A central challenge in applying DiT to long video generation is the ineffective compression of historical context through key-value (KV) caching in standard attention mechanisms. As video sequences extend, this approach results in scalability bottlenecks, since the memory and computation requirements grow proportionally to sequence length. Recent efforts have sought to address this limitation by incorporating recurrent neural networks (RNNs)—including Mamba~\cite{mamba, mamba2, wang2025lingen, po2025longcontextstatespacevideoworld} and Test-Time Training (TTT) approaches~\cite{sun2025learninglearntesttime, zhang2025testtimetrainingright}—within the DiT framework. However, these integrations often suffer from two major issues: (1) chunk-wise autogressive processing, common in TTT-style models, relies heavily on hidden state propagation, causing the model to lose direct access to the dense contextual information present in recent frames and resulting in pixel-level inconsistencies across chunk boundaries; and (2) the introduction of recurrency breaks the parallelizable nature of attention during training, thereby compromising computational efficiency and exacerbating the gap between training and inference procedures.

In this work, we introduce \textbf{Recurrent Autoregressive Diffusion (RAD)}, a unified framework for long-term video generation with global memory and local attention, with the following core designs to address previous challenges:

% We advocate for an explicit frame-wise autoregressive rollout, which better preserves contextual integrity, and introduce a hidden-state prefetch mechanism to recover parallelism in attention computation. Building on these insights, we propose Recurrent Autoregressive Diffusion (RAD)—a unified framework that addresses both the limitations of historical compression and the efficiency constraints imposed by recurrency in long video diffusion.

% Specifically, we incorporate recurrent neural networks (RNNs) into Diffusion Forcing, an autoregressive diffusion framework for long video generation. While the use of RNNs for video generation is well established—with works such as LinGen~\cite{wang2025lingen} demonstrating their effectiveness—the principal advantage of RNNs lies in their fixed-sized hidden states, which remain constant regardless of video length. Although prior studies~\cite{11095233, po2025longcontextstatespacevideoworld, zhang2025testtimetrainingright} have explored integrating RNNs like Mamba~\cite{mamba, mamba2} and TTT~\cite{sun2025learninglearntesttime} into diffusion models, these attempts have encountered challenges such as pixel-level inconsistencies across frame chunks, scalability issues with longer sequences, and an absence of a unified framework for systematic comparison and analysis. 

% In this paper, our main contributions are as follows:
\begin{itemize}
    \item We incorporate an RNN layer along temporal dimension after the attention in each DiT layer for retrieving global memory. Through systematic comparison among different RNN architectures—including LSTM~\cite{6795963}, Mamba2, and TTT—within autoregressive video generation framework, we reveal that LSTM, despite its simplicity, delivers robust performance and often surpasses more recent RNN variants on challenging long video benchmarks.
    \item To preserve local context details, we choose a largely overlapped sliding window attention. Through a comprehensive comparison between \textbf{chunk-wise} and \textbf{frame-wise} autoregressive inference paradigms, we find that frame-wise autoregressive rollout with stride size 1 of sliding window  substantially enhances spatiotemporal consistency by leveraging immediate contextual information, and reduces reliance on persistent hidden state propagation—thereby mitigating pixel-level inconsistencies and improving the fidelity of long video synthesis.
    \item For compute efficiency, we introduce a \textbf{hidden-state prefetch mechanism} to recover parallelism in attention computation, which overcomes the inherent trade-off between recurrency and training parallelism. This mechanism enables fully parallel attention computation during training while retaining the benefits of recurrent historical compression, significantly improving efficiency for large-scale, long-sequence modeling.
    % \item We conduct comprehensive theoretical and empirical comparisons of leading RNN-based approaches—including LSTM~\cite{6795963}, Mamba2, and TTT—across both chunk-wise and frame-wise autoregressive paradigms. Our results demonstrate the substantial advantages of frame-wise generation and reveal that LSTM, despite its simplicity, delivers robust performance and often surpasses more recent RNN variants on challenging long video benchmarks.
\end{itemize}

\section{Related work}

\paragraph{Video Diffusion Model} The remarkable success of diffusion models originated in image synthesis \cite{Rombach_2022_CVPR, ramesh2022hierarchicaltextconditionalimagegeneration} and was later extended to video generation \cite{NEURIPS2022_39235c56, opensora, opensora2, blattmann2023align, hong2022cogvideo, yang2024cogvideox}. Current state-of-the-art video diffusion models typically employ VAEs \cite{kingma2022autoencodingvariationalbayes} to map videos from the pixel level to a latent space, while the model architecture has evolved from the UNet \cite{ronneberger2015unetconvolutionalnetworksbiomedical, chen2024videocrafter2, blattmann2023stable} to diffusion transformer (DiT) \cite{peebles2023scalable}. 
% \zh{maybe add more here if this para is too short}

\paragraph{Video World Model} World models \cite{watter2015embed, ha2018recurrent, hafner2020mastering} predict future states based on the current state and input actions. They hold broad application prospects in fields such as autonomous driving \cite{hu2023gaia, ren2025cosmosdrivedreamsscalablesyntheticdriving}, navigation \cite{bar2024navigation}, and robotic manipulation \cite{wu2024ivideogpt, azzolini2025cosmos, ge2025} and games \cite{valevski2024diffusion, oasis2024, che2024gamegen, guo2025mineworld, yu2025gamefactory}. The capability of video diffusion models to synthesize high-quality videos makes them promising candidates as video world models. Since the proposal of ``video generation models as world simulators" by Sora \cite{videoworldsimulators2024}, a series of foundational world models have emerged—such as Genie2 \cite{parkerholder2024genie2}, Genie3 \cite{genie3}, and Cosmos \cite{nvidia2025cosmosworldfoundationmodel}—demonstrating remarkable video generation quality and interactivity. 
To achieve video world models, recent advancement improves video generation models in multiple aspects: autoregressive inference for extending video durations \cite{chen2024diffusion, huang2025selfforcingbridgingtraintest, cui2025self, Yin_2025_CVPR}, memory mechanism for improving spatiotemporal consistency \cite{chen2025learningworldmodelsinteractive, xiao2025worldmemlongtermconsistentworld, yu2025cam}, physics modeling \cite{kang2024far}, etc.

% However, their reliance on substantial computational resources and large amounts of high-quality data restricts most research efforts to specific settings \cite{}.
% Recent studies \cite{chen2025learningworldmodelsinteractive, zhang2025packinginputframecontext} indicate that one bottleneck in using AR-based video diffusion models as world models is a lack of spatiotemporal consistency. One solution involves incorporating 3D modeling \cite{}, as seen in methods like Marble \cite{marble2025}, yet this approach is constrained by the quality and quantity of available data, making it difficult to scale for large-scale training. Another solution leverages retrieval-based mechanisms \cite{chen2025learningworldmodelsinteractive, xiao2025worldmemlongtermconsistentworld, yu2025cam}, but such methods are limited by reliance on global coordinates and struggle in settings where global coordinates are unavailable. A further approach involves compressing memory using neural networks, typically modeled with RNNs, which we will discuss in detail later.

\paragraph{Diffusion Model with Memory}

While scalable DiT-based video diffusion models have shown strong performance on short clips, applying full attention to long videos incurs prohibitive computational and memory costs. To enable long‑video generation without global attention, additional memory mechanisms are required, which can be classified into \textit{context memory}, \textit{hidden‑state memory}, and \textit{weight memory}. Context memory approaches treat past frames as conditions for autoregressive (AR) prediction. With a limited context window, they either adopt recency‑biased frame selection—as in diffusion forcing \cite{chen2024diffusion, song2025historyguidedvideodiffusion} and self‑forcing \cite{huang2025selfforcingbridgingtraintest, cui2025self}—or employ similarity‑based retrieval, such as VRAG \cite{chen2025learningworldmodelsinteractive} and WorldMem \cite{xiao2025worldmemlongtermconsistentworld}. Hidden‑state memory methods compress historical information into recurrent states, e.g., by inserting Mamba layers before attention \cite{po2025longcontextstatespacevideoworld}. However, the recurrent structure disrupts the temporal parallelism of the original DiT, adding extra computational overhead. Weight memory techniques use inner‑loop losses to update specific weights as memory while sliding over small chunks, as seen in Test‑Time Training (TTT) blocks \cite{11095233, zhang2025testtimetrainingright}. Since adjacent attention windows do not overlap, the updated weights must fully encode all necessary history from prior chunks, imposing high demands on memory capacity for storage and retrieval. In contrast, our approach performs efficient memory compression only for history beyond the context window. It bridges context memory and hidden‑state memory through frame‑wise sliding during autoregressive generation, balancing efficiency and long‑range consistency.
\section{Preliminaries}
% \zh{only RNN and lstm in this section (mamba, ttt can each use one sentence summary and tell why they are rnn), other in appendix; comparison table also needs a lot polishing}

\subsection{Recurrent Neural Networks}\label{sec:preli_rnn}

\paragraph{RNN}
Assuming $x_t \in \mathbb{R}^{D}$ denotes the input of time $t$, 
The general form of a recurrent neural network (RNN) can be described as:
\begin{align*}
    h_t &= f(h_{t-1}, x_t, y_{t-1}; \theta) \\
    y_t &= g(h_t, x_t)
\end{align*}
where $h_t$ is the hidden state at time $t$, $x_t$ is the input, $y_t$ is the output, $f(\cdot)$ is a parameterized nonlinear function (typically involving affine transformation and activation), $\theta$ represents all trainable parameters, and $g(\cdot)$ maps the hidden state to the output.

\paragraph{LSTM}
\begin{align*}
    \begin{bmatrix}
        f_t \\
        i_t \\
        o_t \\
        g_t
    \end{bmatrix}
    &=
    \begin{bmatrix}
        \sigma\\
        \sigma\\
        \sigma\\
        \tanh
    \end{bmatrix}
    W
    \begin{bmatrix}
        y_{t-1} \\
        x_t
    \end{bmatrix}
    \\
    C_t &= f_t \odot C_{t-1} + i_t \odot g_t \\
    y_t &= o_t \odot \tanh(C_t)
\end{align*}

where $f_t, i_t, o_t, g_t$ denote forget gate, input gate, output gate, candidate cell state at time $t$, $\sigma, tanh$ denote sigmoid activation function and hyperbolic tangent activation function, $W$ denote weight and bias matrices for respective gates. The vectors $C_t$ and $y_t$ denote the cell state and output at time $t$, respectively, encapsulating all compressed memory information accumulated up to and including time step $t$.

\paragraph{Mamba}
\begin{align*}
    H_t = AH_{t-1} + BX_{t-1};\quad X_t = CH_t + DX_{t-1}
\end{align*}
where $H_t$ are latent states, and $A,B,C,D$ are linear projections of input $X_t$, i.e., $A_t$ := $\text{Linear}_{\theta_A}$ ($X_t$) and similarly for $B_t$, $C_t$, and $D_t$. This is the mathematical formulation of Mamba for autoregressive tasks. In a multi-layer setting, the output from the previous timestep does not serve as the input for the next timestep, but rather as the input for the next layer. Therefore, it can be expressed as:

\begin{align*}
    H_t = AH_{t-1} + BX_{t};\quad Y_t = CH_t + DX_t
\end{align*}

\paragraph{TTT}
\begin{align*}
    W_t &= W_{t-1}-\eta\nabla \mathcal{L}(W_{t-1};x_t) \\
    y_t &= f(\theta_Qx_t;W_t)
\end{align*}
where the self-supervised loss $\mathcal{L}$ is often defined as $\mathcal{L}(W; x_t) = ||f(\theta_Kx_t;W)-\theta_Vx_t||^2$, $\theta_Q,\theta_K,\theta_V$ are trainable parameters, and $W_t$ is a hidden state matrix at time $t$. TTT-linear learns per-instance weights $W_t \in \mathbb{R}^{d \times d}$, but with often small MLPs or projections. TTT updates global weights per step; typically, there is no recurrent hidden state transfer beyond $\mathcal{L}(W; x_t)$.

\paragraph{Comparison}
As presented in ~\cref{tab:comparison}, we provide a comparison of different RNN types, where $h$ is LSTM hidden size, $n$ is Mamba SSM state size.

\begin{table*}[htbp]
  \centering
  \caption{Comparison of computational complexity between LSTM, Mamba (SSM), and Test-Time Training (TTT-linear with $d\times d$ memory weights) for input tensor $(B,H,W,L,d)$.}
  \label{tab:comparison}
  \begin{resizebox}{\columnwidth}{!}{
  \begin{tabular}{l|c|c|c}
  \hline
  \textbf{Aspect} & \textbf{LSTM} & \textbf{Mamba (SSM)} & \textbf{TTT} \\
  \hline
  Computation (FLOPs) & $O(BHWL(dh + h^2))$ & $O(BHWL\,dn)$ & $O(BHWL(d^2 + C))$ \\
  Parameter count & $O(h(d+h))$ & $O(dn) + O(d^2)$ (if projected) & $O(d^2)$ \\
  Training Memory & $O(BHWLh)$ & $O(BHWL(d+n))$ & $O(BHWd)$ \\
  Inference Memory & $O(BHWh)$ & $O(BHW(d+n))$ & $O(BHWd)$ \\
  Sequence scaling & Linear in $L$ & Linear in $L$ & Linear in $L$ \\
  % % Hidden interaction & Quadratic ($h^2$) & Linear ($dn$) & None \\
  \hline
  \end{tabular}
  }
  \end{resizebox}
\end{table*}

% \paragraph{Other RNNs} Apart from LSTM, Mamba and TTT are two popular recurrent architectures that update hidden states over time. Their key difference lies in how the new state is computed. LSTM updates both its hidden state and cell state using three sources of information—the current input $x_t$, the previous hidden output $y_{t-1}$, and the previous cell state $C_{t-1}$. In contrast, Mamba and TTT follow a simpler RNN-style update that depends only on the current input and the previous hidden state, i.e., $h_t = f(h_{t-1}, x_t)$. More details are provided in Appendix. %Sec.~\ref{sec:mamba_ttt}.

% \paragraph{Comparison}

%  Meanwhile, LSTM differs significantly from the other two RNNs: the update of LSTM's cell states $C_t$ depends on both the input $x_t$ at the current time step and the output $y_{t-1}$ and cell states $C_{t-1}$ from the previous time step. In contrast, Mamba's $H_t$ and TTT's $W_t$ are updated differently—they rely solely on the input $x_t$ at the current time step and the hidden state $H_{t-1}$ and $W_{t-1}$ from the previous time step. This means that the function $h_t = f(h_{t-1}, x_t, y_{t-1}; \theta)$ can be simplified to $h_t = f(h_{t-1}, x_t; \theta)$. This distinction leads to their respective strengths and weaknesses under different training and inference paradigms, which we will analyze in detail in a later section.

\subsection{Video Diffusion model}

\paragraph{Latent Video Diffusion Model}
We adopt a latent video diffusion model~\cite{blattmann2023stable} that first encodes pixel space into a latent representation $\boldsymbol{z} = \mathcal{E}(\boldsymbol{x})$ using a pretrained variational autoencoder (VAE). The forward process gradually adds Gaussian noise to the latent according to a variance schedule $\{\beta_t\}_{t=1}^T$:
\begin{equation}
    q(\boldsymbol{z}_t|\boldsymbol{z}_{t-1}) = \mathcal{N}(\boldsymbol{z}_t; \sqrt{1-\beta_t}\boldsymbol{z}_{t-1}, \beta_t\mathbf{I})
\end{equation}

The model learns to reverse this process by predicting the noise $\boldsymbol{\epsilon}_\theta$ at each step:
\begin{equation}
    \mathcal{L} = \mathbb{E}_{t,\boldsymbol{\epsilon},\boldsymbol{z}}[\|\boldsymbol{\epsilon} - \boldsymbol{\epsilon}_\theta(\boldsymbol{z}_t, t)\|_2^2]
\end{equation}
where $\boldsymbol{z}_t = \sqrt{\bar{\alpha}_t} \boldsymbol{z}_0 + \sqrt{1-\bar{\alpha}_t} \boldsymbol{\epsilon}$ with $\boldsymbol{\epsilon} \sim \mathcal{N}(0,\mathbf{I})$.

At inference time, we can sample new videos by starting from random noise $\boldsymbol{z}_T \sim \mathcal{N}(0,\mathbf{I})$ and iteratively denoising:
\begin{equation}
    \boldsymbol{z}_{t-1} = \frac{1}{\sqrt{\alpha_t}}(\boldsymbol{z}_t - \frac{\beta_t}{\sqrt{1-\bar{\alpha}_t}}\boldsymbol{\epsilon}_\theta(\boldsymbol{z}_t,t)) + \sigma_t\boldsymbol{\epsilon}
\end{equation}
where $\alpha_t = 1-\beta_t$ and $\bar{\alpha}_t = \prod_{s=1}^t \alpha_s$.
The final latent sequence $\boldsymbol{z}_0$ is decoded back to pixel space using the decoder $\mathcal{D}$ to obtain the generated video.

\paragraph{Diffusion Forcing}
% \zh{can move to appendix if no space}
To enable long video generation, we apply the Diffusion Forcing~\cite{chen2024diffusion} technique. During training, we randomly add noise to each frame in the entire input video sequence according to the diffusion schedule: $z^i_t = \sqrt{\bar{\alpha}_t} z^i_0 + \sqrt{1-\bar{\alpha}_t} \epsilon^i, \epsilon^i \sim \mathcal{N}(0,\mathbf{I})$, where $z^i_t$ represents the noised latent of the $i$-th frame, and the training objective for action-conditioned autoregressive video models become:
\begin{align*}
    \mathcal{L}_\text{DF} &= \mathbb{E}_{[t],\boldsymbol{\epsilon},\mathbf{z},a}[\|\boldsymbol{\epsilon} - \boldsymbol{\epsilon}_\theta(\mathbf{z}_{[t]}, [t], \boldsymbol{a})\|_2^2] \\ \boldsymbol{\epsilon}&=\{\epsilon^i\}_{i=1}^L, \mathbf{z}_{[t]}=\{z^i_t\}_{i=1}^L
\end{align*}
where $[t]$ is vector of $L$ timesteps with different $t\in[T]$ for each frame and $\boldsymbol{a}$ is an action sequence $\boldsymbol{a} \in \mathbb{R}^{L \times A}$. The noise prediction model $\boldsymbol{\epsilon}_\theta$ conditioned on both the action sequence $\boldsymbol{a}$ and noised frames $\mathbf{z}_{[t]}$.
\section{Methodology}
\begin{figure*}
    \centering
    \includegraphics[width=0.78\linewidth]{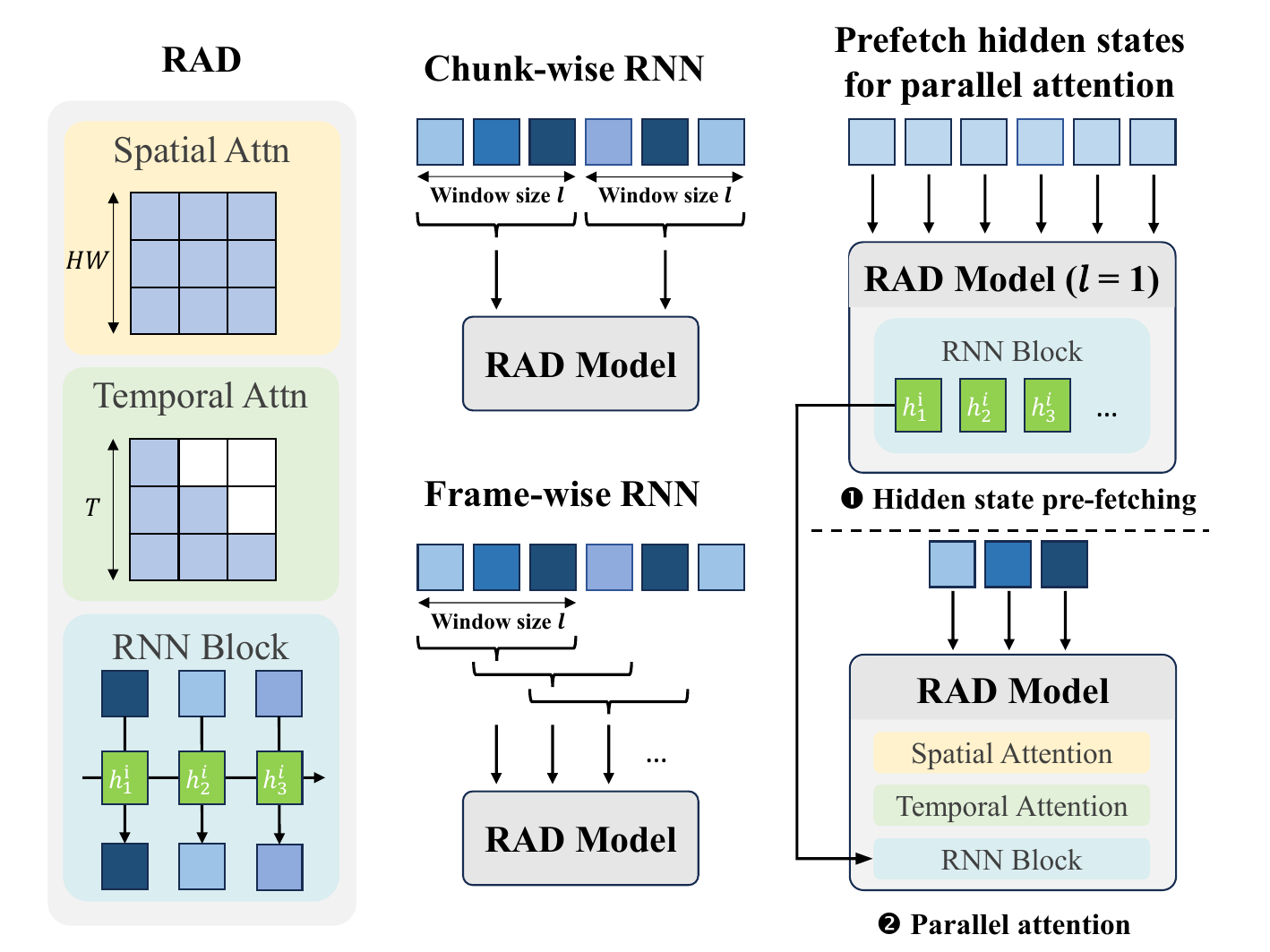}
    \caption{Training paradigm for Recurrent Autoregressive Diffusion with global memory and local attention: The model has three components in each DiT block, including spatial attention, temporal attention and RNN memory block. It supports both chunk-wise and frame-wise autoregressive generation with different attention mechanisms. For efficient training of frame-wise RNN, we (1) pre-fetch the hidden states from clean sample sequence to enable parallel attention computation across entire long sequences, and (2) conduct diffusion model forward in standard manner to get diffusion loss. This improves efficiency and fidelity for long-sequence video modeling. $h^i_j$ is the $i$-th layer hidden state for $j$-th frame.}
    \label{fig:train_paradigm}
    \vspace{-2em}
\end{figure*}

To address the limitations of fixed-size context windows in video diffusion models, we propose the integration of \textbf{global memory with local attention} mechanisms. This approach enables the model to effectively capture long-term dependencies in video sequences while maintaining high fidelity in generated frames.

\subsection{Recurrent Autoregressive Diffusion}
\begin{wrapfigure}[19]{r}{0.56\textwidth}
% \begin{figure}[htbp]
    \vspace{-3em}
    \centering
    \includegraphics[width=\linewidth]{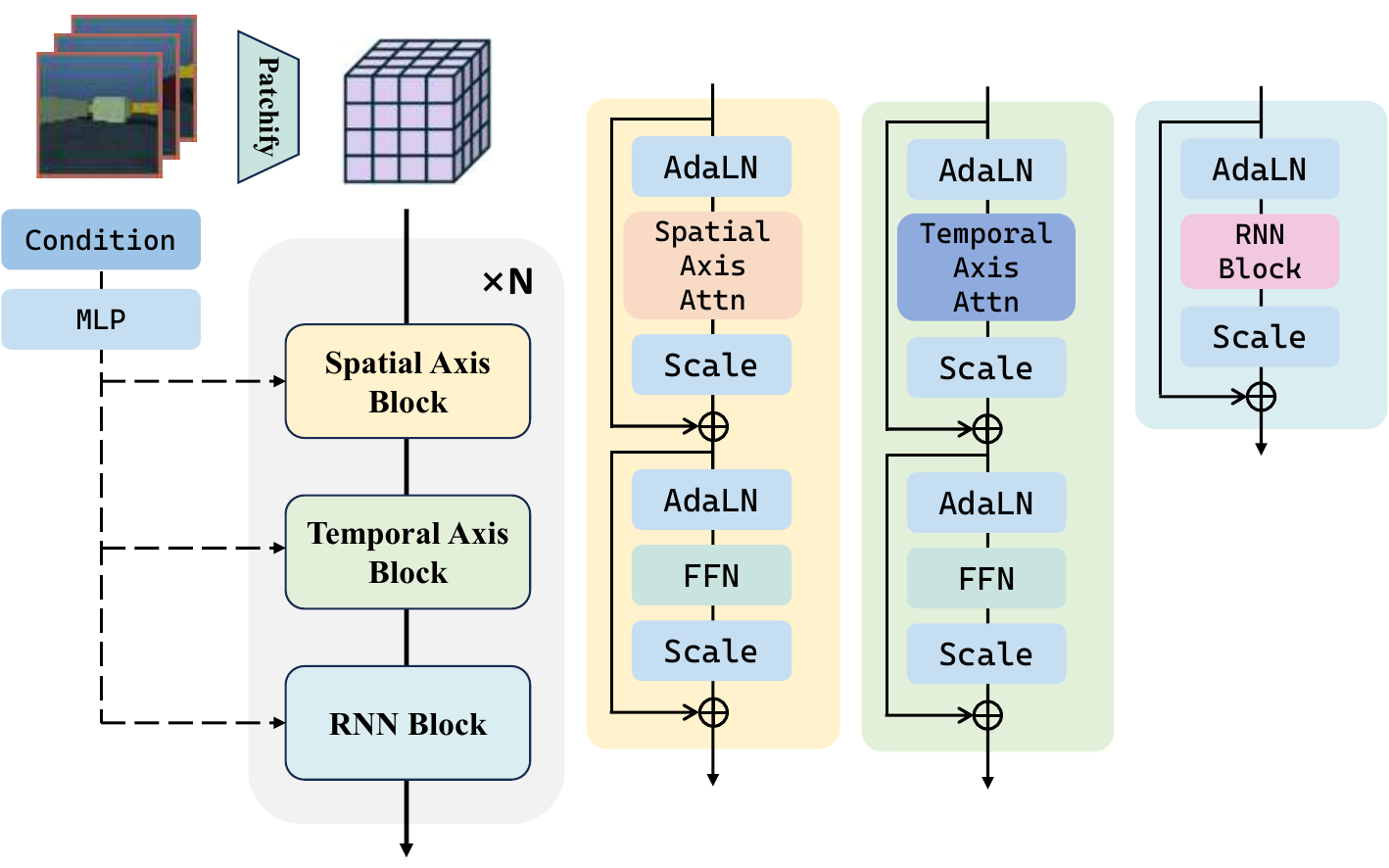}
    \caption{Recurrent Autoregressive Diffusion model architecture: The RAD model consists of $N$ DiT blocks, each of which contains three blocks: a Spatial Axis Block, a Temporal Axis Block, and an RNN Block. The timestep and action condition are processed through an MLP and then used to control each block via adaLN.}
    \label{fig:model_arch}
    \vspace{-1em}
% \end{figure}
\end{wrapfigure}
We introduce the Recurrent Autoregressive Diffusion (RAD) model, which integrates a Recurrent Neural Network (RNN) block into the Diffusion Transformer (DiT) architecture~\cite{peebles2023scalable} to carry global memory information. The overall architecture of the RAD model is illustrated in \cref{fig:model_arch}.
Following the designs by previous work \cite{opensora, oasis2024}, we decompose the attention mechanism in our DiT into two distinct modules: Spatial Axis Attention and Temporal Axis Attention. Each layer of the DiT therefore comprises three primary components—the two attention modules as standard architecture and an additional RNN block.

The RNN block performs retrieval and update operations solely along the temporal axis, meaning computations are carried out on a per-frame basis rather than a per-patch basis. The Rotary Position Embedding (RoPE)~\cite{su2024roformer} is applied in both spatial and temporal dimensions, to enhance the capacity of both attention modules to capture positional dependencies. 
Conditioning information, specifically the timestep and action condition, is incorporated into the RAD model via adaptive Layer Normalization (adaLN). RAD also additionally applies the action conditioning on RNN blocks, apart from the attention modules. This design is verified to effectively enhance the action control and improve video fidelity.

For the RNN block, we compare three alternatives including LSTM, TTT and Mamba's SSM block~\cite{11095233}, and find that LSTM performs the best in our RAD model, shown in experiment \cref{sec:experiment}. 
In the following sections we discuss two key design choices made to optimize the integration of RNNs within the DiT framework: (a). frame-wise autoregression for better context consistency (\cref{sec:chunkwise}) and (b). hidden state pre-fetching for parallel attention computation (\cref{sec:prefetch}).

\subsection{Chunk-wise and Frame-wise Autoregression}\label{sec:chunkwise}

\begin{figure}[hbp]
\vspace{-3em}
    \centering
    \begin{minipage}{0.49\textwidth}
        \centering
        \includegraphics[width=0.7\linewidth]{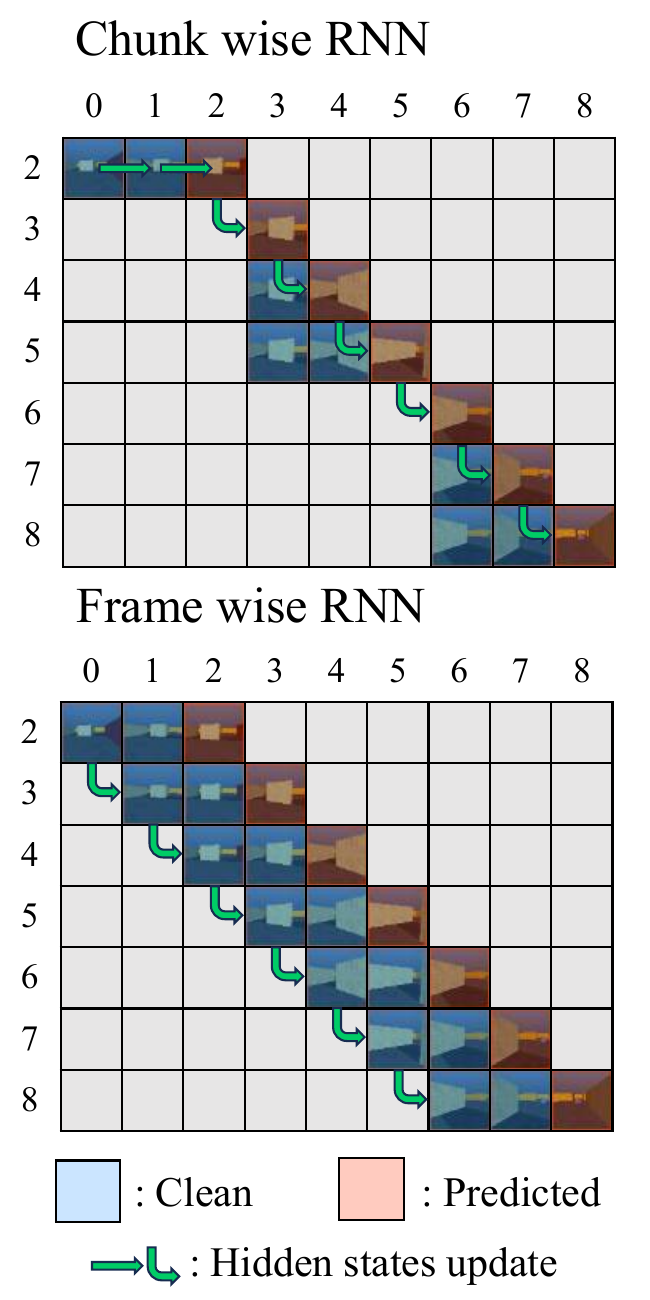}
        \caption{Effective temporal attention maps for chunk-wise and frame-wise RNNs, with chunk size 3 and 2 initial context frames.}
        \label{fig:attn_map}
    \end{minipage}
    \hspace{0.2em}
    \begin{minipage}{0.485\textwidth}
        \vspace{-0.5em}
        \begin{algorithm}[H]
        \scriptsize
        \caption{Inference Pipeline of Frame-wise RAD}
        \label{alg:inference_code}
        \begin{algorithmic}[1]
        \Require Context video $\mathbf{x}_{1:T_c}$, target length $T$, RAD layers of attention and RNN block $\{A_d,R_d\}_{d=1}^D$, window size $C$, denoising steps $K$
        \State $\mathbf{v}_{out}\leftarrow \mathbf{x}_{1:T_c}$
        \State Initialize hidden states $\{h^d\}_{d=1}^D$

        \Statex \textit{\# Prefill hidden states from context}
        \For{$d=1$ to $D$} \hfill \Comment{DiT layer}
            \State $h_0^d\leftarrow\mathbf{0}$
            \For{$t=1$ to $T_c-C+1$}  \hfill \Comment{Frame sequence}
                \State $\mathbf{z}_t^d\leftarrow A_d(\mathbf{x}_t)$
                \State $h_t^d\leftarrow R_d(\mathbf{z}_t^d,h_{t-1}^d)$
            \EndFor
            \State $h^d\leftarrow h_{T_c-C+1}^d$
        \EndFor

        \Statex \textit{\# Sliding window with 1-frame stride}
        \While{$|\mathbf{v}_{out}|<T$} \hfill \Comment{Frame sequence}
            \State Sample $\epsilon\sim\mathcal{N}(0,\mathbf{I})$
            \State $\mathbf{z}\leftarrow \mathrm{cat}(\mathbf{v}_{out}[-C+1:],\epsilon)$
            \For{$k=1$ to $K$} \hfill \Comment{Denoising step}
                \For{$d=1$ to $D$} \hfill \Comment{DiT layer}
                    \State $\mathbf{z}^d\leftarrow A_d(\mathbf{z})$
                    \State $\mathbf{z},\hat{h}^d\leftarrow R_d(\mathbf{z}^d,h^d)$
                \EndFor
                \State $\mathbf{z}\leftarrow \mathrm{DenoiseStep}(\mathbf{z},k)$
            \EndFor
            \State $h^d\leftarrow \hat{h}^d, d=1,\dots,D$ \hfill \Comment{Clean hidden state}
            \State $\mathbf{v}_{out}\leftarrow\mathrm{Append}(\mathbf{v}_{out},\mathbf{z}_{-1})$
        \EndWhile
        \State \Return $\mathbf{v}_{out}$
        \end{algorithmic}
        \end{algorithm}
    \end{minipage}
\end{figure}

% \zh{inference attention map? index for frame?}
In the RAD framework, the RNN block processes temporal information in an autoregressive manner, which can be implemented via two distinct modes: chunk-wise and frame-wise autoregression. Both approaches are seamlessly integrated with the temporal attention mechanisms of the DiT architecture, as illustrated in \cref{fig:train_paradigm} and \cref{fig:attn_map}.

For chunk-wise autoregression, the attention windows are non-overlapping. During training, the input video sequence is divided into chunks based on the model's window size. Within each chunk, local attention is computed, while global temporal dependencies are preserved by propagating the RNN hidden states across chunk boundaries. To ensure that the model operates autoregressively within each chunk, we apply a causal mask to the temporal dimension of the attention mechanism. Importantly, this approach contrasts with existing methods~\cite{11095233, zhang2025testtimetrainingright} that focus on fine-tuning diffusion models trained for full-sequence denoising. Instead, our method ensures both architectural and procedural consistency between training and inference: the model processes data during inference in exactly the same manner as during training, thereby supporting robust temporal generalization and faithful sequence modeling.

For the frame-wise RNN mode, building on Diffusion Forcing \cite{chen2024diffusion}, we employ a frame-by-frame autoregressive generation scheme. This method allows the model to fully leverage the attention mechanism for transmitting pixel-level information across frames. However, this comes at the cost of significant computational overhead when naively applying sliding window-based training. To circumvent this inefficiency, we introduce a Hidden State Pre-fetching strategy, detailed in a subsequent section. During inference, as depicted in \cref{fig:attn_map}, we slide the window one frame at a time, with only the first frame in each window responsible for updating the RNN's hidden state. This procedure directly mirrors the window size of 1 used in the hidden state pre-fetch step at training. Additionally, hidden state updates are performed only at the final step of the DDIM process, ensuring that all memory inputs consist of clean frames—thereby maintaining consistency with the training process and promoting more stable generation quality.

% \subsubsection{ Sliding Temporal Attention}

\subsection{Hidden State Pre-fetch for Parallel Attention}\label{sec:prefetch}
% \zh{this is for frame-wise rnn only}

\begin{figure}[htbp]
    \centering
    \begin{minipage}[t]{0.5\textwidth}
        \vspace{0pt}
        \centering
        \includegraphics[width=0.8\linewidth]{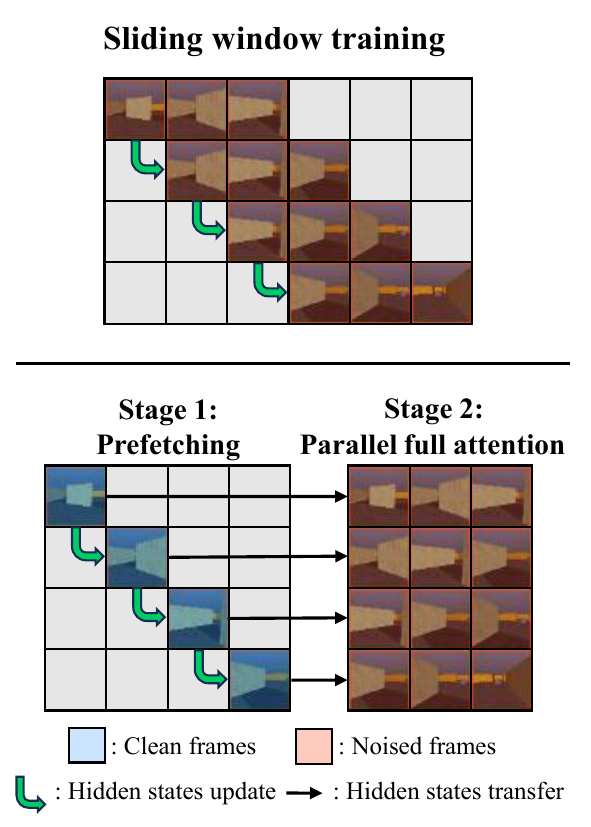}
        \caption{Comparison of standard sliding-window forward and hidden-state prefetching.}
        \label{fig:prefetching}
    \end{minipage}
    \hfill
    \begin{minipage}[t]{0.48\textwidth}
        \vspace{0pt}
        \centering
        \begin{algorithm}[H]
        \scriptsize
        \setstretch{1.18}
        \caption{Frame-wise RAD Training}
        \label{alg:training_code}
        \begin{algorithmic}[1]
        \Require Clean video $\mathbf{x}$, target $\mathbf{v}_{gt}$, sequence length $T$, RAD layers $\{A_d,R_d\}_{d=1}^D$
        \State Initialize hidden bank $\mathbf{H}\leftarrow\emptyset$
        \Statex \textit{\# Prefetch hidden states with 1-frame RAD}
        \For{$d=1$ to $D$} \hfill \Comment{DiT layer}
            \State $h_0^d \leftarrow \mathbf{0}$
            \For{$t=1$ to $T$} \hfill \Comment{Frame sequence}
                \State $\mathbf{z}_t^d \leftarrow A_d(\mathbf{x}_t)$
                \State $h_t^d \leftarrow R_d(\mathbf{z}_t^d,h_{t-1}^d)$
                \State $\mathbf{H}[d,t]\leftarrow h_t^d$ \hfill \Comment{Clean state}
            \EndFor
        \EndFor
        \Statex \textit{\# Train RAD on full sequence}
        \State $\mathbf{z}=\mathbf{x}+\epsilon$, $\epsilon\sim\mathcal{N}(0,\mathbf{I})$
        \For{$d=1$ to $D$} \hfill \Comment{DiT layer}
            \State $\mathbf{z}^d\leftarrow A_d(\mathbf{z})$ \hfill \Comment{Parallel attention}
            \State $\mathbf{z}\leftarrow R_d(\mathbf{z}^d,\mathbf{H}[d,:])$ \hfill \Comment{Parallel RNN}
        \EndFor
        \State \Return $\mathrm{MSE}(\mathbf{z},\mathbf{v}_{gt})$
        \end{algorithmic}
        \end{algorithm}
    \end{minipage}
\end{figure}
Our training strategy for frame-wise RNN is depicted in the right panel of \cref{fig:train_paradigm} and \cref{fig:prefetching}. A well-known limitation of RNNs is their inherent sequential dependence: each timestep’s output is tightly linked to the previous hidden state. This strict temporal recursion inhibits parallelization and considerably slows training, posing a particular challenge for long sequences and large-scale datasets. When RNNs are further integrated with attention mechanisms, the subsequent attention computation needs to follow the same recurrence along the input sequence during training, resulting in excessive computational bottlenecks. It becomes expensive to maintain a fully sequential temporal loop during training.

To mitigate these challenges, we employ a \textbf{hidden state pre-fetching} scheme that partially decouples the RNN from the attention modules. In this framework, the RNN maintains its temporal recurrence, but the attention operations can proceed in parallel, significantly improving training efficiency. Without this technique, a frame-wise sliding with window size $l$ over a sequence length $L$ requires $(L-l+1)$ times sequential attention computation, while parallel attention only conducts once at training in our case.  Another pivotal benefit of this approach is that only clean frames, rather than noised frames from the diffusion process \cite{po2025longcontextstatespacevideoworld}, are used as inputs to RNN memory during prefetching, a design choice that accelerates training convergence (see \cref{sec:clean_mem} for empirical analysis).

Concretely, for frame-wise autoregression, we first apply the RAD model with a window size of 1 across the clean frames of each training sequence. This step independently computes all RNN hidden states $\{h^d_t\}^{d\in [D]}_{t\in[T]}$ across frames in sequence and DiT layers, where each $h^d_t$ aggregates contextual information up to $t$-th frame and $d$-th DiT layer. Assuming the previous hidden state stacks sufficient context information, the pre-fetched hidden states are equivalent to the standard DiT with RNN by sliding-window. However, after pre-fetching all hidden states, the attention modules within each window—despite the sequential RNN update—can now be processed in parallel. By setting the prefetched hidden states for RNN layers at corresponding positions, we can compute the attention modules in RAD with a normal window size as usual to get the diffusion loss for the entire sequence. In our experiments, we do not calculate the diffusion losses for all sub-sequences (length $l$) in given sample (length $L$), but randomly sample partial of them to reduce computation cost.

% \zh{this is equivalent to standard recurrent+attention for due to causal mask along temporal axis?}

% \subsection{Training and Inference Consistency}

% Maintaining consistency between the training and inference stages is crucial. Although our training stage pre-fetches all hidden states, it actually follows the same pattern as the inference stage illustrated in Fig.~\ref{fig:attn_map}. During inference, since we only slide one step at a time, we only allow the first frame to handle the update of the RNN's hidden state. This corresponds to the sliding process with a window size of 1 in the training pre-fetch step. Additionally, we only update the hidden state when reaching the final step of the DDIM process, ensuring that all memory inputs are clean frames, which also remains consistent with the training procedure. \zh{clean frame design, different from mamba?}

% \zh{
% Two key designs need to be highlighted:
% 1. window size 1 forward to get all hidden states first (this works because of causal temporal attention); talk about benefits over iteratively looping windows
% 2. remove lstm loop within each window (use repeated hidden states from beginning of the window) talk about why this avoid overfit.
% }

\section{Experiments}
\label{sec:experiment}

\subsection{Datasets and Evaluation Protocol}\label{sec:dataset}

% \begin{figure}[ht]
%   \centering
%   \begin{subfigure}{0.48\linewidth}
%     \includegraphics[width=\linewidth]{figures/maze_vis/agent_trajectory.png}
%     \caption{Visualization of training data}
%     \label{fig:maze_vis_training}
%   \end{subfigure}
%   \hfill
%   \begin{subfigure}{0.48\linewidth}
%     \includegraphics[width=\linewidth]{figures/maze_vis/eval_vis.png}
%     \caption{Visualization of evaluation data}
%     \label{fig:maze_vis_evaluation}
%   \end{subfigure}
%   \caption{Top view of Memory Maze data}
%   \label{fig:maze_vis}
%   \vspace{-.8cm}
% \end{figure}

\paragraph{Maze Dataset} For our small-scale experimental setting, we employ the Memory Maze dataset~\cite{pasukonis2022memmaze}, which consists of approximately 30000 training videos, each depicting agent navigation within a $15 \times 15$ maze environment and comprising 1000 frames. For the Maze Dataset, all models were trained from scratch during the training process. For evaluation, we collect an additional set of 200 maze videos, with frame counts ranging from 100 to 300. In each sequence, the agent traverses from a designated start position to a target location and then returns along the same path. We define the initial 60\% of frames—corresponding to the outbound trajectory—as the contextual input, while the remaining 40\% serve as the prediction targets for model evaluation. This setup is specifically designed to rigorously test the model's capacity for long-term memory and context utilization.  The aerial visualization of the maze data can be found in the Appendix.

\paragraph{Minecraft Dataset} For large-scale experiments, we utilize the MineRL~\cite{guss2019minerl} to generate 20000 training sequences following the protocols in VRAG \cite{chen2025learningworldmodelsinteractive}. Each video contains 1200 frames. For evaluation, we collect 60 sequences incorporating distinct action patterns, such as rotation in place, which are intended to probe the model's memory capabilities under diverse behaviors. To ensure fair comparison, we first train a standard diffusion forcing model on the entire 20000-sample training set. Subsequently, this pretrained base model is fine-tuned with different RAD-RNN architectures and training paradigms, each for the same 5000 optimization steps. This procedure enables a controlled assessment of architectural and methodological differences under consistent pretraining conditions.

\paragraph{Evaluation Metrics} 
We employ three widely adopted metrics to quantitatively evaluate model performance: Structural Similarity Index (SSIM)~\cite{wang2004image}, which assesses spatial consistency in generated frames; Peak Signal-to-Noise Ratio (PSNR), which measures pixel-level reconstruction fidelity; and Learned Perceptual Image Patch Similarity (LPIPS)~\cite{zhang2018unreasonable}, which evaluates perceptual similarity. Since our evaluation datasets are specially tailored for long-term memory tasks, these metrics that directly compare against ground-truth videos can accurately reflect the capacity of different methods in memory retention and maintaining spatiotemporal consistency.
% Notably, SSIM—when directly comparing generated outputs to ground truth—places a stronger emphasis on the preservation of memory and spatial-temporal consistency, whereas PSNR and LPIPS are more indicative of overall frame quality.

\paragraph{Baselines}
For the baselines, we select Diffusion Forcing~\cite{chen2024diffusion}, Teacher Forcing, and VRAG~\cite{chen2025learningworldmodelsinteractive}. Additionally, TTT-c can be regarded as an implementation of LaCT~\cite{zhang2025testtimetrainingright}. All baselines and our proposed RAD method adopt identical data settings: models are trained from scratch on the Maze dataset, while on the Minecraft dataset, they are fine-tuned starting from a pre-trained model.

\subsection{Maze Results}\label{sec:maze_results}

\begin{table}[ht]
    \centering
    \begin{minipage}{0.48\textwidth}
        \centering        
        \caption{Experiment results on Maze Dataset: chunk-wise (``-c'') and frame-wise (``-f'') autoregressive modes}
        \resizebox{\textwidth}{!}{
        \begin{tabular}{c|c|ccc}
        \toprule
            Model & RNN Type & PSNR $\uparrow$ & SSIM $\uparrow$ & LPIPS $\downarrow$ \\
            \midrule
            DF & None & 14.73 & 0.32 & 0.51 \\
            TF & None & 14.91 & 0.36 & 0.48 \\
            VRAG & None & 15.55 & 0.41 & \textbf{0.43} \\
            \midrule
            \multirow{3}{*}{RAD} & Mamba2-c & 13.80 & 0.31 & 0.58 \\
            & TTT-c & 14.36 & 0.35 & 0.53 \\
            & LSTM-c & \textbf{15.64} & \textbf{0.43} & 0.47 \\
            \midrule
            \multirow{3}{*}{RAD} & Mamba2-f & 15.35 & 0.41 & 0.51 \\
            & TTT-f & 15.50 & 0.41 & 0.52 \\
            & LSTM-f & 15.50 & 0.41 & 0.45 \\
        \bottomrule
        \end{tabular}
        }
        \label{tab:rnn_maze}
    \end{minipage}
    \hfill
    \begin{minipage}{0.48\textwidth}
        \centering
        \caption{Experiment results on Minecraft dataset: chunk-wise (``-c'') and frame-wise (``-f'') autoregressive modes}
        \resizebox{\textwidth}{!}{
        \begin{tabular}{c|c|ccc}
        \toprule
            Model & RNN Type & PSNR $\uparrow$ & SSIM $\uparrow$ & LPIPS $\downarrow$ \\
            \midrule
            DF & None & 15.65 & 0.45 & 0.53 \\
            TF & None & 16.11 & 0.44 & 0.50 \\
            VRAG & None & 16.11 & \textbf{0.49} & 0.50 \\
            \midrule
            \multirow{3}{*}{RAD} & Mamba2-c & 12.72 & 0.33 & 0.63 \\
            & TTT-c & 14.04 & 0.38 & 0.56 \\
            & LSTM-c & 14.24 & 0.39 & 0.55 \\
            \midrule
            \multirow{3}{*}{RAD} & Mamba2-f & 16.70 & 0.46 & 0.47 \\
            & TTT-f & \textbf{16.72} & 0.46 & 0.47 \\
            & LSTM-f & 16.59 & 0.46 & \textbf{0.46} \\
        \bottomrule
        \end{tabular}
        }

        \label{tab:rnn_mc}
    \end{minipage}
\end{table}

\paragraph{RNN Methods}

We compare LSTM, Mamba2, and TTT within the RAD architecture as described in \cref{sec:chunkwise}, trained under identical settings for three epochs. \Cref{tab:rnn_maze} summarizes the results for both chunk-wise (``-c'') and frame-wise (``-f'') autoregressive modes.

In the chunk-wise setting, LSTM delivers the strongest performance across all metrics, improving notably over both the Diffusion Forcing (DF) baseline and the other recurrent variants. Mamba2-c and TTT-c, in contrast, perform worse than the baseline in PSNR and LPIPS, which focus more on image quality. This behavior reflects the burden placed on the recurrent module: without overlapped attention, all pixel-level continuity between chunks must be carried through hidden states or memory weights alone. LSTM benefits structurally from its separation of short-term memory (previous output $y_{t-1}$) and long-term memory (cell state $C_{t-1}$), which aligns well with this requirement. Mamba2 and TTT—designed for global compression rather than fine-grained pixel transport—struggle in comparison.

In the frame-wise setting, all recurrent variants perform similarly. With a sliding step of 1, consecutive frames share an attention window, allowing pixel-level information to propagate directly through attention rather than via hidden states. This eliminates the main failure mode of Mamba2-c and TTT-c, enabling Mamba2-f, TTT-f, and LSTM-f to reach comparable quality with only small metric differences. Correspondingly, LSTM’s structural advantage is diminished, as its explicit short-memory $y_{t-1}$ pathway becomes less necessary when attention already provides strong local continuity. 
Meanwhile, compared with the baseline, the RAD model integrated with the RNN mechanism completely outperforms both Diffusion Forcing and Teacher Forcing. Even when compared to the VRAG method that explicitly models historical frames, the frame-wise RAD model achieves comparable performance.

\paragraph{Frame-wise vs. Chunk-wise Autoregression}

The cross-paradigm comparison in \cref{tab:rnn_maze} highlights how the autoregressive design dictates the relative strength of each recurrent architecture.

In the \textbf{chunk-wise} mode, the absence of local cross-chunk attention forces hidden states to serve as the sole channel for transmitting pixel-level information. This creates excessive demands on its capacity for information storage and retrieval with the memory mechanism: it must encode both global memory and the local scene details needed to maintain visual consistency. Under this constraint, architectural differences become pronounced. Mamba2 and TTT underperform because they are not optimized to shuttle high-frequency information through memory alone, whereas LSTM’s disentanglement of local (via $y_{t-1}$) and global (via $C_{t-1}$) information is highly compatible with the chunk-wise sliding training and inference paradigm. As a result, LSTM-c achieves the strongest performance.

In the \textbf{frame-wise} mode, attention spans all consecutive frames, restoring direct pixel-level communication. Therefore, Hidden states focus mostly on global information, reducing the need for local detail retention. Once this burden is lifted, Mamba2 and TTT improve substantially and converge in performance with LSTM. The recurrent pathway in LSTM becomes partly redundant, which explains the disappearance of its earlier advantage.

Overall, the results indicate that the suitability of an RNN for autoregressive diffusion depends strongly on the temporal granularity of attention: when attention cannot bridge boundaries (chunk-wise), architectures with explicit local–global separation excel; when attention is continuous (frame-wise), architectural differences matter far less. We argue that hidden states could focus more on global information, while high-frequency temporal local information ought to be transmitted primarily through local attention.

\subsection{Minecraft Results}\label{sec:minecraft_results}

\begin{figure*}[htbp]
    \centering
    \includegraphics[width=0.9\linewidth]{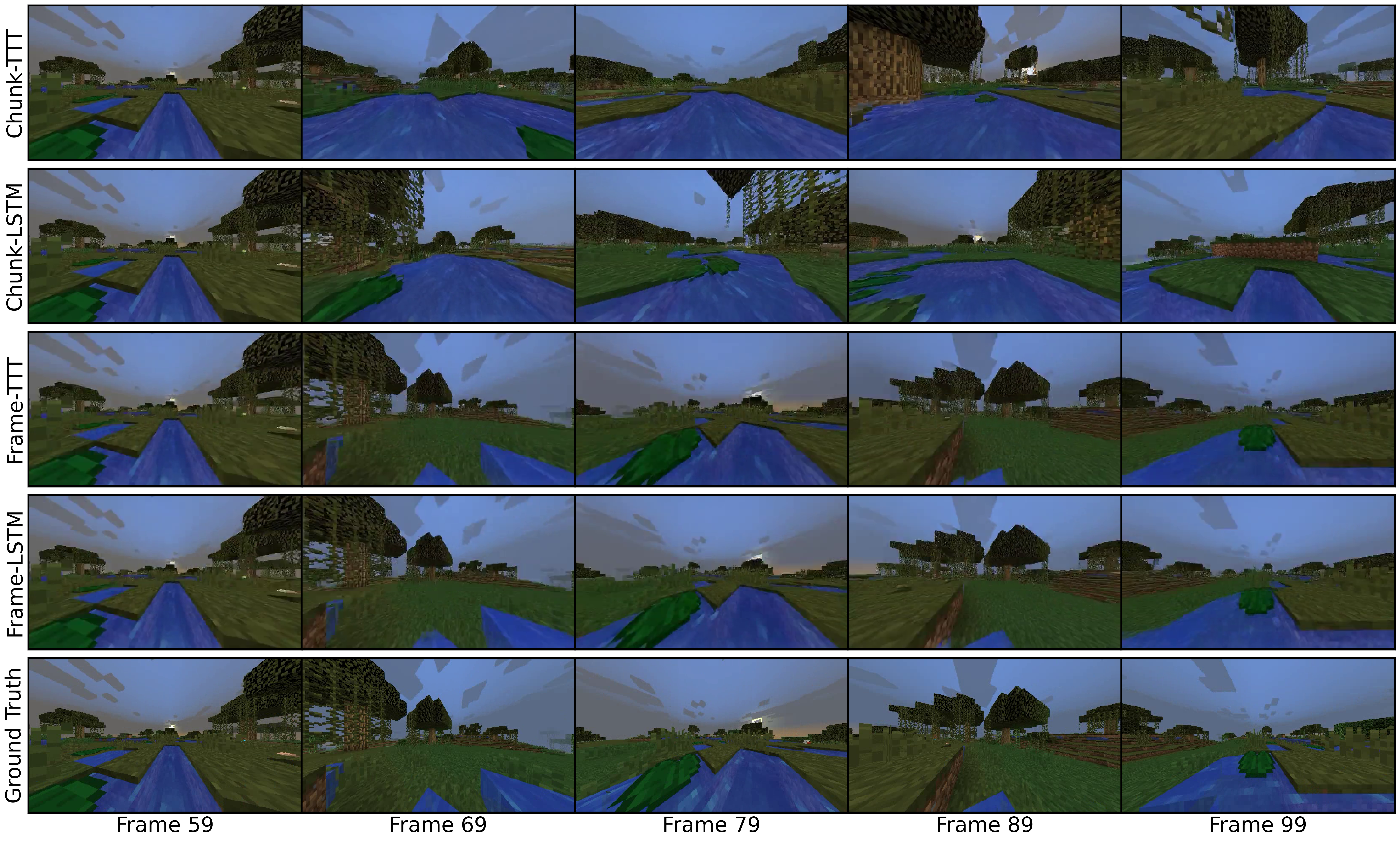}
    \caption{Visualization results on Minecraft dataset. Frame-wise RNN can effectively memorize and reconstruct scenes, maintaining consistency with the ground truth. In contrast, due to the lack of overlap attention, Chunk-wise RNN is unable to reconstruct historical scenes in datasets with higher information density}
    \label{fig:mc_vis}
\end{figure*}

\paragraph{Frame-wise vs. Chunk-wise Autoregression}
\Cref{tab:rnn_mc} presents the results of applying chunk-wise and frame-wise RAD with different RNN variants to the Minecraft dataset. The trends broadly match those observed on the Maze dataset, but with important differences driven by the substantially higher visual and structural complexity of Minecraft videos.

In the \textbf{chunk-wise} setting, LSTM again outperforms Mamba2 and TTT, consistent with its structural ability to balance short-term and long-term memory. However, unlike in the Maze experiments, all RNN variants fall short of the standard Diffusion Forcing baseline. Minecraft scenes contain dense textures, rich geometry, and rapid viewpoint changes, making it difficult for any recurrent hidden state to fully compress and transmit pixel-level information across chunk boundaries. As a result, the architectural limitations of relying solely on hidden states for inter-chunk communication become much more pronounced, leading to degraded reconstruction quality and weaker scene consistency.

In the \textbf{frame-wise} setting, the RAD models with three different RNN types exhibit similar performance, significantly surpassing the chunk-wise counterparts, because local visual information can propagate directly through attention, removing the need to encode high-frequency details in the hidden state. Under this more favorable regime, all three RNN types achieve significantly better performance than their chunk-wise counterparts.

\begin{wrapfigure}[]{r}{0.35\textwidth}
    % \centering
    \vspace{-1em}
    \includegraphics[width=\linewidth]{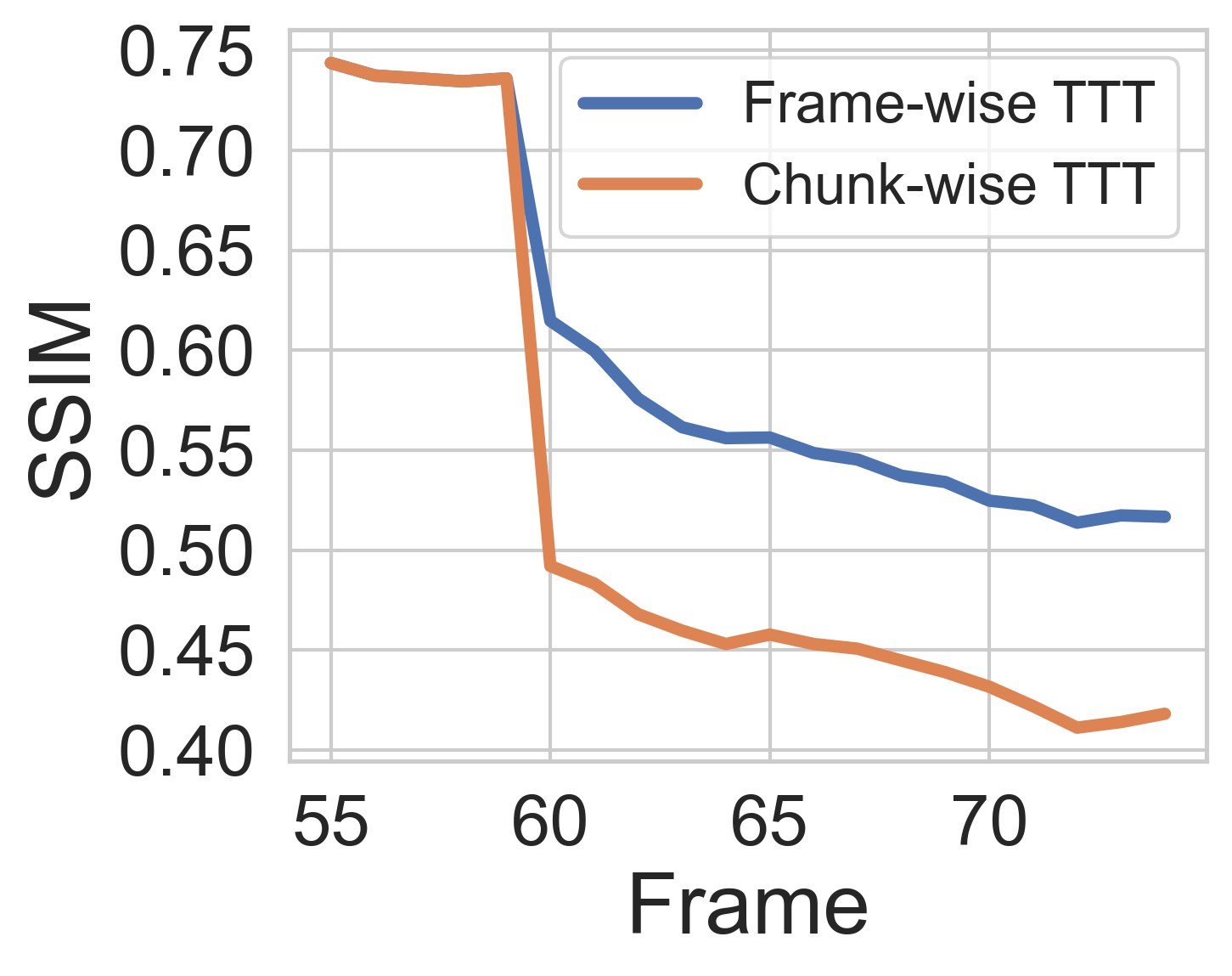}
    \caption{Comparison of chunk-wise TTT and frame-wise TTT on SSIM metric}
    \label{fig:mc_compare}
    \vspace{-2em}
\end{wrapfigure}

Qualitative results in \cref{fig:mc_vis} highlight this contrast. Frame-wise TTT and frame-wise LSTM produce videos that maintain strong memory and stay aligned with ground truth across long horizons. Conversely, chunk-wise TTT and chunk-wise LSTM exhibit clear failures in memorizing scene layout and object configuration, reinforcing the difficulty of relying solely on hidden states to carry rich Minecraft-level detail across chunks.

Overall, these experiments further support the conclusion that the granularity of the autoregressive window plays a critical role. Chunk-wise autoregression imposes an unrealistic compression burden for visually complex environments, while frame-wise autoregression leverages attention to maintain local fidelity, allowing all RNN architectures to operate on more global signals and achieve substantially better results.

To provide a more detailed comparison between chunk-wise and frame-wise RNN approaches, we present the SSIM value curves as a function of frame index in \cref{fig:mc_compare}. Notably, both methods exhibit a decrease in performance at the 60-th frame, corresponding to the first frame predicted by the model. However, this drop is substantially more pronounced for the chunk-wise RNN, underscoring its limited capacity to effectively convey pixel-level information across chunk boundaries. 

\subsection{VBench Results}

\begin{table}[htbp]
\vspace{-1.5em}
    \centering
    \caption{VBench evaluation results for different autoregressive model variants.}
    \resizebox{\columnwidth}{!}{
        \begin{tabular}{l|ccccc|ccccc}
        \hline
 & \multicolumn{5}{c|}{Maze Dataset} & \multicolumn{5}{c}{Minecraft Dataset} \\
\hline
Metrics & \makecell{Background\\Consistency} & \makecell{Temporal \\Flickering} & \makecell{Motion \\Smoothness} & \makecell{Aesthetic \\Quality} & \makecell{Imaging \\Quality} & \makecell{Background\\Consistency} & \makecell{Temporal \\Flickering} & \makecell{Motion \\Smoothness} & \makecell{Aesthetic \\Quality} & \makecell{Imaging \\Quality} \\
% DF & 0.9123 & 0.9308 & 0.6966 & 0.274 & 0.5258 \\
 \hline
LSTM-c & \textbf{91.29} & \textbf{93.49} & 71.4 & \textbf{26.29} & 51.44 & 97.42 & 93.96 & 95.12 & 57.58 & \textbf{69.34}  \\
Mamba-c & 91.07 & 93.07 & 69.98 & 25.51 & 50.7 & 97.27 & 93.79 & 94.96 & 57.22 & 69.27 \\
TTT-c & 90.89 & 92.21 & 70.71 & 26 & \textbf{51.6} & \textbf{97.44} & 93.91 & 95.06 & 57.55 & 69.15 \\
LSTM-f & 91.16 & 93.01 & \textbf{71.48} & 26.11 & 50.94 & 97.38 & \textbf{94.21} & \textbf{95.3} & 55.71 & 66.72 \\ 
Mamba-f & 90.78 & 92.97 & 70.78 & 25.76 & 50.94 & 97.37 & 94.17 & 95.26 & \textbf{55.73} & 66.4 \\ 
TTT-f & 90.99 & 93.2 & 70.87 & 25.82 & 51.08 & 97.38 & 94.14 & 95.25 & 55.64 & 66.26 \\ 
\hline
        \end{tabular}
    }
    \label{tab:vbench}
\end{table}
We add VBench~\cite{huang2023vbench} evaluation results beyond just pixel similarity. As shown in \cref{tab:vbench}, the results on VBench demonstrate similar outcomes to those in \cref{sec:maze_results} and \cref{sec:minecraft_results}. The LSTM-based RNN achieves better performance, enabling the generation of higher-quality videos.

\subsection{Computational Resource Analysis}

\begin{wrapfigure}[11]{r}{0.53\textwidth}
    \vspace{-3em}
  \centering
  \captionof{table}{Comparison of computational resource requirements among different RNN block types. ``Time'' denotes the duration of a single training step for an RAD model (same DiT) with the specified RNN type, while ``Params'' indicates the total number of parameters in the RNN block.}
  \resizebox{\linewidth}{!}{
    \begin{tabular}{c|ccc}
    \toprule
        RNN Type & GPU Mem (GB) & Time (s/step) & Params (M) \\
        \midrule
        LSTM & 10.0 & 3.7 & 195 \\
        Mamba2 & 11.1 & 3.6 & 153\\
        TTT & 16.0 & 12.9 & 118 \\
    \bottomrule
    \end{tabular}
    }
  \label{tab:computational}
\end{wrapfigure}

Directly comparing the computational efficiency of different RNN architectures is inherently challenging, as their implementations and underlying optimizations can vary substantially. For example, LSTM benefits from extensive low-level optimizations, such as those provided by cuDNN, whereas TTT currently lacks such specialized enhancement. Despite these differences, we report a summary of computational resource costs for reference. All reported results are obtained from experiments on the Minecraft dataset, conducted using 8 NVIDIA L40 GPUs and a batch size of 3.

As presented in \cref{tab:computational}, although the LSTM architecture features the highest parameter count, its GPU memory consumption and training time per step are comparable to those of Mamba2. These discrepancies in parameterization reflect intrinsic differences in model design rather than unfair experimental conditions, permitting a reasonable assessment.

Meanwhile, to quantify the computational overhead introduced by our method, we also theoretically calculate the Flops cost of each component as in Tab.~\ref{tab:flops}.

\begin{table}[htbp]
\vspace{-1em}
\centering
\caption{Theoretical FLOPs analysis of one forward pass in the proposed two-stage pipeline.}
\label{tab:flops}
\small
\setlength{\tabcolsep}{5pt}
\begin{tabular}{lclc}
\toprule
Component & Formula & FLOPs & Fraction \\
\midrule
Stage 1 (RNN) & {\scriptsize $L(BN)Tn_{\rm layers}\,8D^2$} & $5.1\times10^{13}$ & 35.7\% \\
Stage 1 (Attention) & {\scriptsize $L\!\left(4(BT)ND^2+2(BT)N^2D\right)$} & $2.5\times10^{13}$ & 17.5\% \\
Stage 2 (Diffusion) & 
    {\scriptsize $\begin{aligned}
    &L\!\left(4(B_2W)ND^2+2(B_2W)N^2D\right)\\
    &+L\!\left(4(B_2N)WD^2+2(B_2N)W^2D\right)
    \end{aligned}$}
& $6.7\times10^{13}$ & 46.9\% \\

\bottomrule
\end{tabular}
\vspace{-1em}
\end{table}

% \begin{align*}
%     % \mathrm{M}_{RNN,step}&\approx 4(D_{in}D + D^2)=8D^2 \\
%     \mathrm{Flops}_{\mathrm{stage_1,rnn}}&=L\cdot (BN)\cdot T\cdot n_{\mathrm{layers}}\cdot 8D^2\approx 5.1\times 10^{13} \\
%     \mathrm{Flops}_{\mathrm{stage1,attn}}&= L\Big(4(BT)ND^2 + 2(BT)N^2D\Big)
%     \approx 2.5\times 10^{13} \\
%     \mathrm{Flops}_{\mathrm{stage2}}&= L\Big(4(B_2W)ND^2 + 2(B_2W)N^2D\Big) \\
%     &+ L\Big(4(B_2N)WD^2 + 2(B_2N)W^2D\Big) \approx6.7\times 10^{13}
% \end{align*}

Where $L$ denotes the number of DiT layers, $N$ is the token length, $W$ represents the window size, and $T$ is the number of video frames. We can observe that the additional computational overhead introduced by prefetching mainly comes from the attention computation in Stage 1, which accounts for less than 20\% of the total computational cost. Diffusion Forcing baseline has equivalent computational cost to only the Stage 2 diffusion forward pass.
\section{Ablation study}
\subsection{Noise Level of Memory Frames}\label{sec:clean_mem}

\begin{table}[htbp]
  \centering
  \small
  \begin{minipage}[t]{0.42\textwidth}
    \vspace{0pt}
    \centering
    \caption{Evaluation results for LSTM-f with clean and noised frame memory.}
    \label{tab:lnr}
    \vspace{0.6em}
    \resizebox{0.59\textwidth}{!}{
    \begin{tabular}{c|cc}
      \toprule
      Metrics & noised & clean \\
      \midrule
      PSNR $\uparrow$ & 14.70 & \textbf{15.30} \\
      SSIM $\uparrow$ & 0.38 & \textbf{0.41} \\
      LPIPS $\downarrow$ & 0.52 & \textbf{0.50} \\
      \bottomrule
    \end{tabular}
    }
  \end{minipage}
  \hfill
  \begin{minipage}[t]{0.56\textwidth}
    \vspace{0pt}
    \centering
    \caption{Strided chunk-wise autoregression for LSTM on Maze dataset.}
    \label{tab:partial_overlap}
    \vspace{0.6em}
    \resizebox{\textwidth}{!}{
    \begin{tabular}{l|c|ccc}
      \toprule
      RNN Type & Stride & PSNR $\uparrow$ & SSIM $\uparrow$ & LPIPS $\downarrow$ \\
      \midrule
      LSTM-c & 20 & \textbf{15.64} & \textbf{0.43} & 0.47 \\
      LSTM-f & 1 & 15.50 & 0.41 & \textbf{0.45} \\
      LSTM-partial-overlap & 10 & 15.06 & 0.40 & 0.55 \\
      \bottomrule
    \end{tabular}
    }
  \end{minipage}
\end{table}

We conducted an ablation study to evaluate the impact of noise levels applied to memory frames that are fed into hidden states of RNN during training. Our findings indicate that introducing higher noise levels makes model optimization more challenging, leading to significantly higher training loss and degraded evaluation metrics, as illustrated in \cref{tab:lnr}. These results highlight the importance of our design choice: all clean frames used for memory are pre-processed in the pre-fetching stage, slid by a DiT model with a window size of 1 and decoupled from the denoising stage. 

This approach shares some similarities with the strategy of Po et al. \cite{po2025longcontextstatespacevideoworld}, where the initial $n$ frames of each training sequence are left clean and excluded from the diffusion loss calculation. However, our method is more comprehensive, where all frames passed into the RNN hidden states remain clean, rather than solely the initial portion. Moreover, our approach does not interfere with the computation of the diffusion loss, further enhancing training efficiency.

\subsection{Strided Chunk-wise Autoregression}\label{sec:abla_stride}
We further investigate the partially overlap attention window in chunk-wise RNN, by keeping the same model size and training data as Sec.~\ref{sec:maze_results}, while setting the stride of the sliding window to 10 frames (consistent during both training and inference). In other words, for frame-wise RNN, the stride is 1; for chunk-wise RNN, the stride equals the window size (20). The experimental results, as shown in Tab~\ref{tab:partial_overlap}, indicate that this method does not show advantage compared to both frame-wise RNN and chunk-wise RNN.

% \subsection{Design of RNN block}
% \subsection{Action Condition Design}\label{sec:abla_cond}
% We performed an ablation study to investigate the influence of RNN block design, focusing on how conditional information is incorporated. In the standard approach, timestep and action condition information are introduced to the DiT block via AdaLN, which may limit the RNN block’s ability to utilize these signals. To address this, we experiment with passing the action condition through a linear projection and concatenate it directly with the output of the attention block over the hidden dimension, using this combination as the input to the RNN block.

% As shown in \cref{tab:rnn_chunk_no_comb}, this modification leads to a notable improvement for LSTM, but produces only limited gains for TTT and Mamba architectures. Remarkably, LSTM continues to outperform both TTT and Mamba even without this integration, highlighting its inherent superiority for modeling sequential information. 

\section{Discussion and Conclusion}
% The proposed RAD framework integrates RNN blocks into DiT modules for memory enhancement of autoregressive video generation. Experiments show superior performances of LSTM over other RNN variants, indicating its advantages. We also demonstrates the improvement by switching chunk-wise to frame-wise autoregression for long video fidelity.

In this work, we present RAD, a unified framework for long-term video generation that augments the DiT architecture with RNN memory blocks. By systematically comparing LSTM, Mamba2, and TTT within the RAD framework, we find that LSTM achieves strong and robust performance for challenging long-range video synthesis tasks. We further analyze the importance of memory update paradigms, and demonstrate that switching from chunk-wise to explicit frame-wise autoregressive generation with overlapping context windows dramatically improves spatiotemporal consistency and video fidelity, with the RAD framework striking a good balance between local context details and global memory. Although the performance advantage is verified, we found that there are still limits of the current method: the additional RNN blocks introduce noticeable computational overhead compared to standard diffusion models, which leads to increased training and inference time. Meanwhile, the constant-sized hidden state of the RNN has a theoretical storage upper bound, which may restrict the model performance under ultra-long context scenarios.
% To address the computational challenges of training recurrent models, we introduce a hidden-state prefetch mechanism, which enables efficient parallel attention computation while retaining the benefits of sequential memory. Experiments on Memory Maze and Minecraft benchmarks verify that our approach substantially improves long video generation quality. We hope RAD will provide a foundation for future research in scalable, interactive video world models.

% \section*{Acknowledgments}
% ---- Bibliography ----
%
% BibTeX users should specify bibliography style 'splncs04'.
% References will then be sorted and formatted in the correct style.
%
\bibliographystyle{splncs04}
\bibliography{main}
\end{document}